\pdfoutput=1
\documentclass[11pt]{article}

\usepackage[]{EACL2023}

\usepackage{times}
\usepackage{latexsym}
\usepackage{color}
\usepackage{longtable}
\usepackage{comment}
\usepackage{hyperref}
\usepackage{multirow}
\usepackage{graphicx}
\usepackage{tabularx}
\usepackage{stfloats}
\usepackage{todonotes}

\usepackage[T1]{fontenc}
\usepackage[utf8]{inputenc}

\usepackage{microtype}

\usepackage{inconsolata}

\title{ChatGPT: Beginning of an End of Manual Linguistic Data Annotation? \\ Use Case of Automatic Genre Identification}

\author{Taja Kuzman \and Igor Mozetič\\
  Jožef Stefan Institute, Slovenia \\
  \texttt{taja.kuzman@ijs.si} \\
  \texttt{igor.mozetic@ijs.si} \\ \And
  Nikola Ljubešić \\
  Jožef Stefan Institute, Slovenia \\
  Center za jezikovne vire in tehnologije \\ Univerze v Ljubljani, Slovenia\\
  \texttt{nikola.ljubesic@ijs.si} \\}

\begin{document}
\maketitle
\begin{abstract}
ChatGPT has shown strong capabilities in natural language generation tasks, which naturally leads researchers to explore where its abilities end. In this paper, we examine whether ChatGPT can be used for zero-shot text classification, more specifically, automatic genre identification. We compare ChatGPT with a multilingual XLM-RoBERTa language model that was fine-tuned on datasets, manually annotated with genres. The models are compared on test sets in two languages: English and Slovenian. Results show that ChatGPT outperforms the fine-tuned model when applied to the dataset which was not seen before by either of the models. Even when applied on Slovenian language as an under-resourced language, ChatGPT's performance is no worse than when applied to English. However, if the model is fully prompted in Slovenian, the performance drops significantly, showing the current limitations of ChatGPT usage on smaller languages. The presented results lead us to questioning whether this is the beginning of an end of laborious manual annotation campaigns even for smaller languages, such as Slovenian.
\end{abstract}

\section{Introduction}

ChatGPT has shown strong capabilities as a dialogue system, providing clearer and more helpful answers than humans \citep{guo2023close}. It remains unclear whether its performance on text categorization tasks, more specifically on automatic genre identification, can be compared to the existing large language models (LLMs), such as the XLM-RoBERTa model \citep{conneau2020unsupervised}, fine-tuned to the task. Despite the fact that the ChatGPT model was made available to the public only a few months ago, some studies, analyzing the potential of the model for numerous natural language processing (NLP) tasks, have already been published. \citet{qin2023chatgpt} analyzed its zero-shot performance on reasoning tasks%(arithmetic, commonsense, symbolic and logical reasoning)
, natural language inference, dialogue, question answering, summarization, named-entity recognition and sentiment analysis. The results showed that ChatGPT was outperformed by the fine-tuned large language models at most tasks. It achieved significantly worse results in symbolic reasoning and named-entity recognition. Similarly, \citet{zhong2023can} compared ChatGPT with fine-tuned language models BERT \citep{DBLP:journals/corr/abs-1810-04805} and RoBERTa \citep{liu2019roberta} on the GLUE benchmark \citep{wang2018glue}, consisting of sentiment analysis, linguistic acceptability, paraphrase, textual similarity, natural language
inference, and question answering. The overall results showed that ChatGPT performed comparably to the BERT model, while it was outperformed by the RoBERTa model. In contrast, when \citet{zhang2022would} analyzed ChatGPT's performance on stance detection, it achieved state-of-the-art (SOTA) results on this task. The ChatGPT model was also shown to achieve competitive performance in machine translation of high-resource languages, while it is lacking behind the SOTA models on low-resource languages based on the WMT22 \citep{kocmi-etal-2022-findings} dataset \citep{hendy2023good}.

In addition to comparing the model to fine-tuned LLM models, some studies compared the model's performance with manual annotation. When \citet{huang2023chatgpt} examined the performance of ChatGPT on categorizing implicit hate speech and providing natural language explanations for the implicit hate speech detection, the results showed great potential of ChatGPT for this and similar tasks. The authors report that the results are promising, as the model correctly identified 80\% of implicit hateful tweets. What is more, its explanations were evaluated to be better than those, provided by humans. Based on this, the authors see a ``great potential of ChatGPT as a data annotation tool''. %The authors also that the explanations, provided by the ChatGPT, were perceived by annotators as clearer than the human-generated explanations. However, the authors also warn that the result show that the model's explanations can reinforce human's perception and warn that its convincing capability might mislead the user if its chosen label is wrong.

In this paper, we examine ChatGPT's performance on automatic genre identification, a text classification task, where manual annotation has been repeatedly shown to be very hard for humans \citep{egbert2015developing, zu2004genre, suchomel2020genre}. In addition to providing first insights into the performance of ChatGPT on this task, this is also one of the first studies of its performance on a language, other than English. We compare ChatGPT's zero-shot performance with the X-GENRE classifier -- a multilingual XLM-RoBERTa Transformer-based language model, fine-tuned on manually-annotated genre datasets. The models are compared on two test sets, English EN-GINCO dataset and Slovenian GINCO dataset \citep{kuzman-rupnik-ljubei:2022:LREC}, in three scenarios: English prompt with English text, English prompt with Slovenian text and Slovenian prompt with Slovenian text. The experiments show impressive performance of ChatGPT on this task, outperforming the fine-tuned LLM on English test set. In addition, despite Slovenian being an under-resourced language, ChatGPT's performance on this language is no worse than on English, provided that the prompt is in English instead of Slovenian. So much time and effort went into manual annotation of datasets in numerous languages to fine-tune the models for genre identification, which does not seem to be necessary anymore. Is this a point for the NLP community to stop and ask ourselves: ``Have large annotation campaigns became redundant? Can we use ChatGPT to annotate data for research purposes instead?''

The paper is organized as follows. In Section \ref{sec:related}, we introduce the task of automatic genre identification. We present genre-annotated datasets in Section \ref{sec:dataset}, on which we test the ChatGPT model and the fine-tuned LLM, presented in Section \ref{sec:models}. We discuss the predictions of ChatGPT in Section \ref{sec:chatgpt-predictions} and compare it with the fine-tuned model in Section \ref{sec:comparison}. Finally, in Section \ref{sec:conclusion}, we conclude the paper with discussion of the main findings and suggestions for further work. 

\section{Automatic Genre Identification}
\label{sec:related}

Automatic genre identification is a text classification task which focuses on categorizing texts into genre categories, such as \textit{News}, \textit{Legal}, \textit{Promotion}. The genre categories are traditionally defined based on the author's purpose, and the function and conventional form of the text \citep{orlikowski1994genre}. Ever since the emergence of the world wide web, researchers both from the information retrieval as well as the corpora creation and curation field have approached this task with the aim of creating systems for automatically identifying genres of texts. This information would allow the users of information retrieval tools to obtain more relevant hits to their queries. In addition, the advent of the web allowed for a fast collection of corpora based on automatic methods. Since a challenge with this approach is that the corpora creators and users do not know what texts are included in the resulting web corpus~\citep{baroni2009wacky}, annotating these large collections with genre information provides valuable insight into the content and quality of a web corpus. Moreover, genre information was shown to be a beneficial signal for numerous NLP tasks, such as language processing \citep{giesbrecht2009part, muller2021genre}, machine translation \citep{van2018evaluation}, and automatic summarization \citep{stewart2009genre}.

In contrast to topic detection, genre categories cannot be categorized solely based on lexical information, such as keywords. The models need to find a higher pattern in texts, often based on textual or syntactic characteristics, unrelated to the topic of the document. In addition, classification of genres was shown to be a hard task because texts can be more or less prototypical examples of their genre classes, can show signals of multiple classes or lack signals of any genre \citep{sharoff2021genre, zu2004genre}. That is why this text categorization task is very challenging for non-neural methods which were shown to be too dataset-dependent and were not capable of generalizing to unseen datasets \citep{sharoff2010web}.

Then a breakthrough happened with the advent of deep neural Transformer-based language models. By using the BERT-like language models that are pre-trained on massive amounts of texts, and by fine-tuning them on genre identification task, recent works \citep{Kuzman_Ljubesic_Pollak_2022, ronnqvist2021multilingual,repo2021beyond} showed that the models are capable of identifying genres even on unseen datasets and languages. In addition, they achieve good results already when trained on only 1,000 texts. However, the models still need manually-annotated texts, which is a time-consuming and expensive task.

%\citet{qin2023chatgpt} analyze the ChatGPT zero-shot performance on 7 natural language processing tasks and 20 datasets. They analyse the model's performance on various reasoning tasks (arithmetic, commonsense, symbolic and logical reasoning), natural language inference, dialogue, question-answering, summarization, named-entity recognition and sentiment analysis. The study compares two dialogue models, the ChatGPT model and the GPT-3.5 model. The results showed that ChatGPT is outperformed by the fine-tuned models in the most tasks, with more than 30 points lower scores in symbolic reasoning, on the COPA dataset as part of the commonsense task and named entity recognition.

%\citet{huang2023chatgpt} examine the performance of ChatGPT on categorizing implicit hate speech and providing natural language explanations for the implicit hate speech detection. The results are promising, as the model correctly identified 80\% of implicit hateful tweets which the authors see as a ``great potential of ChatGPT as a data annotation tool''. The authors also that the explanations, provided by the ChatGPT, were perceived by annotators as clearer than the human-generated explanations. However, the authors also warn that the result show that the model's explanations can reinforce human's perception and warn that its convincing capability might mislead the user if its chosen label is wrong.

\begin{table*}
\centering
\begin{tabular}{lccc}
\hline
\textbf{Labels}  & \textbf{EN-GINCO}& \textbf{GINCO}&\textbf{X-GENRE training} \\ 
\hline
Information/Explanation & 25\% & 24\%  & 17\%   \\
Promotion               & 22\%  & 17\% & 16\%   \\
Opinion/Argumentation   & 18\%   & 11\% & 14\%  \\
News                    & 18\%  & 29\%  & 19\% \\
Other                   & 6\%   & 7\% & 4\%   \\
Forum                   & 6\%  & 5\% & 8\%     \\
Instruction             & 5\%  & 5\% & 12\%   \\
Legal             & 0\%  & 1\% & 4\%  \\
Prose/Lyrical            & 0\%  & 1\% & 6\%   \\
\hline  
\end{tabular}
\caption{Label distribution in the English test set (EN-GINCO), the Slovenian test set (GINCO) and the dataset, used for training of the X-GENRE classifier.}
\label{tab:label-distribution}
\end{table*}

\section{Genre-annotated datasets}
\label{sec:dataset}

To evaluate the models' performance on English and Slovenian texts, we use random samples from two manually-annotated datasets: EN-GINCO and GINCO. We test the models on 100 instances from each dataset. As the X-GENRE classifier was trained on the training split of GINCO, we sample the instances to test the models from the test split. In contrast, the EN-GINCO dataset was not part of the training data, used for fine-tuning the X-GENRE model.

The GINCO dataset \citep{kuzman-rupnik-ljubei:2022:LREC} consists of Slovenian web texts, originating from two Slovenian web corpora, the slWaC 2.0 corpus \citep{erjavec2014slwac} from 2014 and the MaCoCu-sl 1.0 corpus\footnote{\url{http://hdl.handle.net/11356/1517}} from 2021. The EN-GINCO dataset is a sample of English texts from the English web corpus enTenTen20\footnote{\url{https://www.sketchengine.eu/ententen-english-corpus}} \citep{jakubivcek2013tenten}. The EN-GINCO dataset has not been published online yet, so it is impossible that ChatGPT would have seen it before. The GINCO dataset was published in 2021\footnote{The dataset is available at \url{http://hdl.handle.net/11356/1467}}.

The instances were manually annotated with 24 genre categories from the GINCO schema by two annotators with linguistic background and following detailed guidelines for genre annotation\footnote{\url{https://tajakuzman.github.io/GINCO-Genre-Annotation-Guidelines/}} (see \citet{kuzman-rupnik-ljubei:2022:LREC} for more details on the annotation procedure). In case of disagreement, the annotators discussed the instance on which they did not agree, and jointly decided on the final label. The inter-annotator agreement in GINCO, calculated before the discussions of the annotators, reached nominal Krippendorff's alpha of 0.71, which is above the acceptable threshold of 0.67, defined by \citet{krippendorff2018content}. However, the alpha is well below 0.8 -- which represents good reliability of annotated data -- which further confirms the difficulty of manual annotation for this task, addressed in multiple studies \citep{suchomel2020genre, egbert2015developing}.

\subsection{Genre schema}

In our experiments, we use the X-GENRE genre schema, a generalization of various schemata applied on different datasets, that is, the CORE \citep{egbert2015developing}, FTD \citep{sharoff2018functional} and GINCO \citep{kuzman-rupnik-ljubei:2022:LREC} dataset. The motivation behind the schema is two-fold -- it is more user-friendly than any of the specific schemata in any of the available training datasets, and it allows for merging training data from different datasets, resulting in a more robust model. For this purpose, we mapped the original labels in both test sets, which were annotated with a schema with higher granularity, to the X-GENRE schema. The final schema consists of 9 labels: \textit{Information/Explanation, Instruction, Legal, News, Opinion/Argumentation, Promotion, Prose/Lyrical, Forum} and \textit{Other} (for more details, see the definitions of the labels in Table \ref{tab:genre-description} in the Appendix).

As shown in Table \ref{tab:label-distribution}, each test set is unbalanced, with 4 more frequent categories representing more than 80\% of instances in each test set, and five less common categories, representing 7\% or less of the set each. Two categories from the genre schema appear only in the Slovenian test data: \textit{Legal} and \textit{Prose/Lyrical}. From the comparison of distributions, we can also see that the Slovenian test set has less \textit{Promotion} and \textit{Opinion/Argumentation}, while it has more \textit{News} texts. As label distribution can impact the performance of the classifier, we also add into the comparison the distribution of genre classes in the dataset, on which the X-GENRE classifier was trained on. As shown in Table \ref{tab:label-distribution}, the classes that are less frequent in the test sets are more represented in the training data, and some of the most frequent categories from the test sets, such as \textit{Information/Explanation} and \textit{News}, represent a smaller percentage of the dataset, compared to the test sets.

\section{Models}
\label{sec:models}

\subsection{ChatGPT}

ChatGPT is a large language model, provided by the OpenAI, that was fine-tuned on the GPT-3.5 model \citep{openai}. The model was optimized for dialogue based on the Reinforcement Learning with Human Feedback \citep{christiano2017deep}, that is, a method that uses human feedback to optimize the model's answers. Firstly, the language model is fine-tuned on a dataset of prompts and human-generated answers. Secondly, the model is asked to produce multiple answers to the prompts from the dataset. Then humans evaluate which answer is the best, and their feedback is used to learn a reward function. More specifically, the reward
model is trained on the dataset with evaluated answers to predict which output the labelers would prefer, and the trained model is used as a reward function. Finally, the reward function is optimized with reinforcement learning using the proximal policy optimization algorithm \citep{ouyang2022training}. We use the ChatGPT Feb 13 version and perform the experiments in the period from February 24th to March 2nd 2023.

\subsection{X-GENRE}

We compare the ChatGPT model with a massively multilingual base-sized XLM-RoBERTa Transformer-based model \citep{conneau2020unsupervised}, fine-tuned on genre-annotated datasets, hereinafter referred to as the X-GENRE classifier. The classifier was fine-tuned on around 1,700 instances from three datasets, manually annotated with genre labels: English CORE \citep{egbert2015developing}, English FTD \citep{sharoff2018functional} and Slovenian GINCO \citep{kuzman-rupnik-ljubei:2022:LREC} dataset. Each dataset has its own set of categories, which were then mapped into a joint schema -- the X-GENRE schema. We opted for using multiple datasets instead of just one in order to assure better generalization of the model to unseen datasets and languages, and to avoid the model being too dataset-dependent, which was shown to be an issue in the past \citep{sharoff2010web}.

When we compared the X-GENRE model with models, trained only on one of the datasets, mentioned above, and each using their original schemata, the X-GENRE model outperformed all single-dataset-based models, reaching micro and macro F1 scores between 0.79 and 0.80 in in-dataset testing scenario. We also compared the X-GENRE classifier with models, trained on a single genre dataset, to which the categories from the joint (X-GENRE) schema were mapped, to be able to compare the cross-dataset capabilities of models. The results showed the X-GENRE model to outperform single-dataset-based models on the EN-GINCO dataset, with the single-dataset-based models reaching 0.49 to 0.65 in micro F1 and 0.42 to 0.54 in macro F1\footnote{The experiments and results are documented at \url{https://github.com/TajaKuzman/Genre-Datasets-Comparison/tree/main/Creation-of-classifiers-and-cross-prediction}.}.

The model is freely available at the Hugging Face repository\footnote{\url{https://huggingface.co/classla/xlm-roberta-base-multilingual-text-genre-classifier}}.

\section{ChatGPT prompt and genre predictions}
\label{sec:chatgpt-predictions}

To classify the texts according to genre categories with ChatGPT, we fed it prompts using the OpenAI platform and manually extracted the categories and explanations from its answers. In the prompt, we defined the main criteria for defining genres and which categories can the model choose from. We also asked the model to provide an explanation of its choice, and we provided the text to be classified. The prompt was repeated with each text. We used the genre classes that are used by the X-GENRE classifier to be able to compare the two models.

Example of the English prompt:
\textit{Please classify the following text according to genre (defined by function of the text, author's purpose and form of the text) and explain your decision. You can choose from the following classes: News, Legal, Promotion, Opinion/Argumentation, Instruction, Information/Explanation, Prose/Lyrical, Forum, Other. The text to classify: Shower pods install in no time… <p> 1. Prepare the floor with the waste and the water supply pipes. <p> 2. Attach shower equipment to the shower pod shell running flexible tails (H\&C or just C) down back. <p> 3. Move unit into position connecting water supplies on the way and the waste outlet trap. <p> 4. Having secured the shower pod shell to the building structure doors may now be fitted.}

Example of the ChatGPT's answer:

\textit{Genre: Instruction}

\textit{Explanation: The text provides step-by-step instructions for installing shower pods. It is structured in a sequential manner, detailing the process of preparing the floor, attaching shower equipment, moving the unit into position, and securing the shower pod shell. The language used is imperative, giving clear and concise commands to the reader. There is no attempt to persuade, promote, or argue for any particular point of view. Instead, the purpose of the text is to instruct the reader on how to install shower pods. Therefore, the text can be classified as an instruction manual or guide.}

\begin{table*}[b]
\centering
\begin{tabular}{cllrrr}
\hline
\multicolumn{1}{l}{\textbf{Test set}} & \textbf{Prompt}     & \multicolumn{1}{l}{\textbf{Model}} & \textbf{Micro F1} & \textbf{Macro F1} & \textbf{Accuracy} \\
\hline
\multirow{2}{*}{EN-GINCO}            & \multirow{2}{*}{EN} & ChatGPT                           & 0.74              & 0.66              & 0.72              \\
                                     &                     & X-GENRE                            & 0.67              & 0.61              & 0.67           \\  
\hline
\multirow{2}{*}{GINCO}            & \multirow{2}{*}{EN}       & ChatGPT        & 0.75              & 0.64              & 0.75         \\
                                     &                     & X-GENRE        & 0.91              & 0.91              & 0.91           \\  
\hline
\multirow{2}{*}{GINCO}            & \multirow{2}{*}{SL}       & ChatGPT        & 0.68 &     0.56 &    0.68          \\
                                     &                     & X-GENRE        & 0.91              & 0.91              & 0.91           \\ 
\hline
\end{tabular}
\caption{Comparison of ChatGPT and fine-tuned X-GENRE model on the two test sets and with prompts in two languages.}
\label{tab:comparison}
\end{table*}

In most cases, the answers followed the form of the example, consisting of the label in one line, and the explanation in the next paragraph or multiple paragraphs. However, in some cases, the label was a part of the explanation.

The Slovenian prompt was a translation of the English prompt, where also the names of the classes were translated into Slovenian. When we used the Slovenian prompt, ChatGPT's response with the label and explanation was in Slovenian as well. However, in one instance, the model started the answer in Slovenian but finished the explanation in English. An instance of Slovenian prompt and ChatGPT's response in Slovenian is shown in Subsection \ref{sec:slovene-prompt} in the Appendix.

In a few cases, ChatGPT's output consisted of two classes instead of one. If one of the classes matched the true label, we used this class for evaluation. In addition, in a few cases, ChatGPT answered with a novel label that was not a part of the genre class set, such as \textit{Review} and \textit{Condolence}. We experimented with asking it to correct its decision and choose a label from the closed set. As can be seen from the dialogue in Subsection \ref{sec:chatgpt-correction} in the Appendix, the model apologized and then predicted the correct label.

\section{Comparison with the fine-tuned model}
\label{sec:comparison}

We compare the two models in three scenarios: on English test set (EN-GINCO) with English prompt, on Slovenian test set (GINCO) with English prompt, and on Slovenian test set (GINCO) with Slovenian prompt. In the latter two scenarios, only the language of the prompt is different, while the text instances to classify are the same. The results are shown in Table \ref{tab:comparison}.

When both models are tested on the EN-GINCO test set, surprisingly, ChatGPT outperforms the X-GENRE classifier by 5--7 points in micro F1, macro F1 and accuracy, achieving micro F1 of 0.74, macro F1 of 0.66 and accuracy of 0.72. Considering the fact that the X-GENRE classifier was fine-tuned on more than 1,700 manually-annotated texts, which required laborious annotation campaigns, while ChatGPT was not explicitly trained for this task, these results are astonishing.

In contrast, when the models are tested on the Slovenian test set, the fine-tuned X-GENRE classifier significantly outperforms ChatGPT, reaching micro F1, macro F1 and accuracy scores of 0.91. However, we should note that the X-GENRE model was trained on the training portion of the GINCO dataset, while the test split of the same dataset is used for testing in this scenario. Thus, while the results of the X-GENRE model on the EN-GINCO test set reflect its cross-dataset performance, the results on the GINCO test set show its in-dataset performance, which explains its significantly higher results on this dataset.

A more interesting result is the comparison of ChatGPT's performance on English texts and Slovenian texts. We can see from Table \ref{tab:comparison} that ChatGPT's performance on Slovenian is comparable to English, despite the fact that Slovenian is considerably less present in the training data, used for pre-training and fine-tuning the ChatGPT model, since it is an under-resourced language.% and the fact that it represented 1\% of training data for a similar OpenAI model, GPT-3\footnote{\url{https://github.com/openai/gpt-3/blob/master/dataset\_statistics/languages\_by\_word\_count.csv}}.

In contrast, the performance of ChatGPT drops once we use the Slovenian prompt instead of the English prompt. As shown in Table \ref{tab:comparison}, on Slovenian prompt, ChatGPT achieves 7 to 8 points lower micro F1, macro F1 and accuracy scores than on English prompt, while the text instances are the same. This shows that while the language of the text to be predicted does not have a large impact, the language of the prompt does have an effect on the performance of ChatGPT on automatic genre identification.

\begin{figure}
	\centering		\includegraphics[width=\linewidth]{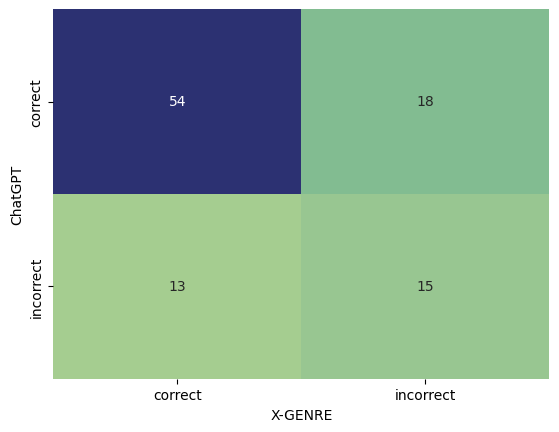}
	\caption{Comparison of differences in correct and incorrect predictions between ChatGPT and X-GENRE.}
	\label{fig:chatgpt-x-genre}
\end{figure}

To analyze the differences between the two models further, we examined the cases where the models did not agree in their predictions. We performed the analysis on the results of the experiments on the EN-GINCO test set, where an English prompt and English text were used. As we can see from Figure \ref{fig:chatgpt-x-genre}, both models mostly correctly predicted genre categories, with ChatGPT being accurate in 72\% of instances and X-GENRE in 67\% of instances. In 69\% the responses of the models match, which means that they were either both correct or both incorrect. 18 instances were correctly predicted by ChatGPT, but incorrectly predicted by X-GENRE, while the X-GENRE model correctly predicted 13 instances to which ChatGPT assigned wrong labels. Out of all instances where ChatGPT is correct, while X-GENRE is incorrect, 39\% are of \textit{Information/Explanation} genre and 33\% belong to \textit{Promotion}. In contrast, instances where X-GENRE is correct and ChatGPT is incorrect mostly belong to \textit{Opinion/Argumentation} genre (38\%) and \textit{Information/Explanation} (23\%). We can conclude from this analysis that, as expected given the different nature of the two models, the outputs of the two models are quite complementary, and that a merge of the outputs could be considered in specific use cases.

\section{Conclusions}
\label{sec:conclusion}

Up to now, supervised machine learning was the method which achieves the highest results in most NLP tasks. It consists of manually annotating thousands of texts with labels, preferably by multiple annotators, on which the machine learning systems are to be trained and tested. However, manual annotation requires a lot of time, human effort and money. What is more, despite great efforts to assure reliable annotation, the annotation campaigns often result in poor inter-annotator agreement, and with that, poor reliability of the annotated data. This reflects how difficult these tasks are even for humans. Results from some previous studies \citep{huang2023chatgpt} and our experiments hint that ChatGPT could substitute large manual annotation campaigns, which would significantly change the workflow of researchers in NLP.

In this paper, we analyze ChatGPT's zero-shot performance on the task of automatic genre identification. We compare it with the X-GENRE classifier, a Transformer-based language model that was fine-tuned on more than 1,700 texts, manually-annotated with genres in three annotation campaigns and two languages. Surprisingly, the results show that when the models are compared on a dataset, on which neither of them was trained, ChatGPT outperforms the X-GENRE classifier. Until now, the paradigm in text classification was that the better and the more manually-annotated data, the better the model. However, these results might hint at a new era for text categorization, where only minimal manual annotation for test sets would be necessary.

Despite the very promising results on the English dataset, our expectation was that the ChatGPT model would not be up to the task on texts in other languages, especially in Slovenian as an under-resourced language. However, when the model was applied both on English and Slovenian genre-annotated dataset, the results show stable performance regardless of the language. In contrast, once the model is also prompted in the under-resourced language, the results start to deteriorate. While this result shows the obvious limitation of the current ChatGPT model, it does not mean much for the data-annotation-oriented usage of the model in the research community, as prompting the model in English surely does not present any challenge for researchers.

However, the difference in results, impacted by the prompt, indicates a need for more detailed research on prompts in this task. A possible direction for further work would be to experiment with more advanced prompting techniques to find the structure which enables us to get the most out of the ChatGPT model, such as manual few-shot chain-of-thought prompting \citep{wei2022chain} which was shown to significantly improve ChatGPT's performance on classification tasks \citep{zhong2023can}. In addition, as multiple genre schemata exist of various granularity, it would be interesting to compare how well does ChatGPT predict genres based on different schemata, e.g., the CORE schema \citep{egbert2015developing}, FTD schema \citep{sharoff2018functional}, GINCO schema \citep{kuzman-rupnik-ljubei:2022:LREC} and so on, compared to the X-GENRE schema, used in this research.

In addition, we plan to extend the comparison in multiple ways. Firstly, while the present experiments were performed on small test sets of 100 instances, recently, an official ChatGPT API has been made available, which will allow for much quicker classification experiments. Thus, we plan to extend the comparison to bigger test sets in more languages and from more genre-annotated datasets. Secondly, like some of previous studies \citep{qin2023chatgpt, hendy2023good}, we plan to extend the comparison to other GPT models, such as the GPT-3.5 model, as well as more BERT-like models, fine-tuned on different genre-annotated datasets.

Furthermore, as web texts often do not undergo any review, automatic genre identification was shown to be a hard task also due to hybrid texts, that is, texts having characteristics of multiple genres. In addition, each document from web corpora consists of all text extracted from one web page. Consequently, another challenge for genre identification are multi-part documents, where a document consists of two or more separate texts of separate genres, such as a document from a news portal that consists of a news article and user comments. In our preliminary experiments, ChatGPT assigned multiple classes to some texts, and was also shown to be able to recognize that a document consists of different parts, and assign genre classes to each part of the document. Thus, as further work, we plan to explore what are the capabilities of ChatGPT for multi-label classification and for genre identification on spans of documents.

Finally, our results have sparked questions whether extensive manual annotation is still needed for such text classification tasks. In the future, we plan to explore the impact of the sizes of manually-annotated datasets to the performance of fine-tuned models, compared to the zero-shot performance of ChatGPT. In addition, as using ChatGPT and similar models for annotation of very large text collections is computationally expensive, we might experiment with using large language models to annotate the training data as an alternative to manual annotation, and then fine-tune base-sized models for annotation of large quantities of data.

%- idea: use large language models to annotate training data and then fine-tune base-sized models to use them for annotation of large quantities of data
%initiatives of building such powerful open and freely-available model. It took a lot of time to analyse 100 instances - multiple hours of manually inserting the prompts as queries and extracting the label and explanation. In contrast, classification with X-GENRE classifier took 3 minutes on 100 instances, with no post-processing to extract the labels necessary. We needed multiple hours to annotate 100 instances, because after a certain number of instances, the ChatGPT did not allow us to continue with queries for an hour or more with the notification ``Too many requests in 1 hour. Try again later.". Sometimes, producing the explanation took more than 1 minute. Sometimes, the dialogue just stopped with the notification ``Something went wrong, please try reloading the conversation.'' or we were thrown out of the platform with the error notification ``Application error: a client-side exception has occurred (see the browser console for more information)''. The answers are not always a clear label, sometime the label is a part of the explanation. \\

\section*{Acknowledgements}
This work has received funding from the European Union's Connecting Europe Facility 2014-2020 - CEF Telecom, under Grant Agreement No. INEA/CEF/ICT/A2020/2278341. This communication reflects only the author's view. The Agency is not responsible for any use that may be made of the information it contains. This work was also funded by the Slovenian Research Agency within the Slovenian-Flemish bilateral basic research project ``Linguistic landscape of hate speech on social media'' (N06-0099 and FWO-G070619N, 2019–2023)
and the research programme ``Language resources and technologies for Slovene'' (P6-0411).

\bibliography{main}
\bibliographystyle{acl_natbib}

\appendix

\section{Appendix}
\label{sec:appendix}

\subsection{Slovenian prompt and answer}
\label{sec:slovene-prompt}

\textbf{Slovenian prompt:}

\textit{Prosimo, določi naslednjemu besedilu, kakšnega žanra je (ki ga definiramo na podlagi funkcije besedila, avtorjevega namena in oblike besedila) in utemelji svojo odločitev. Izbiraš lahko med naslednjimi kategorijami: Novica, Pravno besedilo, Promocijsko besedilo, Mnenje/Argumentativno besedilo, Navodila, Informativno/razlagalno besedilo, Proza/poezija, Forum, Drugo. Besedilo, ki mu določi žanr: Opis: Onesnaževala iz cestnega prometa so eden izmed glavnih virov, ki vplivajo na koncentracije delcev PM10 v mestih, ter negativno vplivajo na človekovo zdravje. V diplomski nalogi obravnavamo korelacije med kakovostjo zunanjega zraka z delci PM10 in prometnimi podatki iz avtomatskih števnih mest v Mariboru. Po naših ugotovitvah korelacij med omenjenimi parametri ni, zato smo se odločili za postavitev novega merilnega mesta v središču Maribora. Z novim merilnim mestom bomo dobili bolj izpopolnjeno merilno mrežo v MOM, boljše informacije o kakovosti zraka v središču mesta hkrati pa možnost boljše prepoznave zveze med prometom in koncentracijami delcev PM10. Najdeno v: ključnih besedah Povzetek najdenega:... Onesnaževala iz cestnega prometa so eden izmed glavnih... Ključne besede: promet, Maribor, onesnaževala, prašni delci PM10 Objavljeno: 27.09.2011 Ogledov: 719 Prenosov: 73 Polno besedilo (20,99 MB)}

\textbf{Chat GPT's response:}

\textit{Žanr tega besedila je Informativno/razlagalno besedilo. Besedilo je namenjeno predstavitvi rezultatov diplomske naloge, ki raziskuje korelacije med kakovostjo zunanjega zraka in prometom v Mariboru ter ponuja predlog za postavitev novega merilnega mesta v središču mesta. Besedilo vsebuje znanstvene podatke in razlage, ki so namenjene informiranju bralca o problematiki onesnaževanja zraka v mestu. Besedilo ni namenjeno prepričevanju bralca, ampak zgolj predstavljanju dejstev in ugotovitev raziskave.}

\begin{comment}
\begin{figure}[!h]
	\centering		\includegraphics[width=\linewidth]{Example-of-Slovene-prompt.png}
	\caption{An example of the Slovenian prompt and ChatGPT's genre prediction and explanation in Slovenian language.}
	\label{fig:slovene-prompt}
\end{figure}
\end{comment}

\subsection{ChatGPT correction}
\label{sec:chatgpt-correction}

In the following example, ChatGPT chose the label outside of the class set, and after an additional prompt, corrected itself, ruling out all classes and correctly identifying the text as \textit{Other}.

\textbf{Prompt:}

\textit{Please classify the following text according to genre (defined by function of the text, author's purpose and form of the text) and explain your decision. You can choose from the following classes: News, Legal, Promotion, Opinion/Argumentation, Instruction, Information/Explanation, Prose/Lyrical, Forum, Other. The text to classify: Condolence: Dear George,Charlene,Michael and family; I am so sorry to see you have lost your beloved Yolande. She was a wonderful person to know and converse with...always with a warm friendly smile and a twinkle in her eye(:>) Marion will sure miss playing dominos with her at the Camp. May your fond memories give you strength and hope knowing you will be reunited again. Jesus cares and is praying for you. May you sense His presence and peace in your time of sorrow. Love in Christ Jan and Bill}

\textbf{ChatGPT's answer:}

\textit{The genre of the given text is "Condolence," which is a type of personal communication expressing sympathy and sorrow upon someone's loss or bereavement. The purpose of the text is to offer condolences and comfort to the family of the deceased. The text's form is informal and personal, written in the form of a letter to the bereaved family, expressing the author's emotions and feelings about the loss. Therefore, the text is not a news item, legal document, promotional material, instructional guide, or forum post. It is not a piece of prose or lyrical writing, nor is it an argumentative opinion piece. Therefore, it does not fit any of the other genres listed.}

\textbf{Researcher's response:}

\textit{"Condolence" is not in the list of allowed classes. Please choose from the following classes: News, Legal, Promotion, Opinion/Argumentation, Instruction, Information/Explanation, Prose/Lyrical, Forum, Other, and explain your decision.}

\textbf{ChatGPT's response:}

\textit{Apologies for the confusion. I would classify the given text under the "Other" category. The text is a personal message expressing condolences and offering comfort to the family of the deceased. It is not a formal document, news item, legal statement, promotional material, instructional guide, or forum post. It is not an opinion or argumentative piece nor is it purely informative or explanatory. The text's purpose is to convey sympathy, share memories, and offer emotional support, making it a form of personal communication. Therefore, the "Other" category seems to be the most appropriate classification for this text.}

\begin{comment}
\begin{figure}[!h]
	\centering		\includegraphics[width=\linewidth]{ChatGPT-correction.png}
	\caption{An example of a dialogue where ChatGPT chose the label outside of the class set, and after an additional prompt, corrected itself, ruling out all classes and correctly identifying the text as \textit{Other}.}
	\label{fig:chatgpt-correction}
\end{figure}
\end{comment}

\newpage
\subsection{Genre categories}

\begin{table*}[b]
\centering
\begin{tabularx}{\textwidth}{m{0.22\linewidth}m{0.40\linewidth}m{0.25\linewidth}}
\hline
\textbf{Label}                   & \textbf{Description}                                                                                                                                                                                                                                                                                                                                                                                                                                                  & \textbf{Examples}                                                                                                                                                                                                                                \\
\hline
Information/Explanation & An objective text that describes or presents an event, a person, a thing, a concept etc. Its main purpose is to inform the reader about something.                                                                                                                                                                                                       & research article, encyclopedia article, product specification, course materials, biographical story/history \\
\hline
Instruction             & An objective text which instructs the readers on how to do something.                                                                                                                                                      & how-to texts, recipes, technical support                                                                                                                                                                                                \\
\hline
Legal                   & An objective formal text that contains legal terms and is clearly structured.                                                                                                                                                                                   & small print, software license, terms and conditions, contracts, law, copyright notices                                                                                                            \\
\hline
News                    & An objective or subjective text which reports on an event recent at the time of writing or coming in the near future.                                                                                                                                                                                            & news report, sports report, police report, announcement                                                                                                                                                         \\
\hline
Opinion/Argumentation   & A subjective text in which the authors convey their opinion or narrate their experience. It includes promotion of an ideology and other non-commercial causes.                                                                     & review, blog, editorial, letter to editor, persuasive article or essay, political propaganda                                        \\
\hline
Promotion               & A subjective text intended to sell or promote an event, product, or service. It addresses the readers, often trying to convince them to participate in something or buy something.  & advertisement, e-shops, promotion of an accommodation, promotion of company's services, invitation to an event                                                                                                 \\
\hline
Prose/Lyrical           & A literary text that consists of paragraphs or verses. A literary text is deemed to have no other practical purpose than to give pleasure to the reader. Often the author pays attention to the aesthetic appearance of the text. It can be considered as art.                                                                                                                                                                                               & lyrics, poem, prayer, joke, novel, short story                                               \\
\hline
Forum & A text in which people discuss a certain topic in form of comments. & discussion forum, reader/viewer responses, QA forum \\
\hline
Other & A text which does not fall under any other genre category. & \\
\hline
\end{tabularx}
\caption{Descriptions of genre labels, with examples.}
\label{tab:genre-description}
\end{table*}

\end{document}